# Do Different Deep Metric Learning Losses Lead to Similar Learned Features?

Konstantin Kobs    Michael Steininger    Andrzej Dulny    Andreas Hotho

University of Würzburg

Germany

{kobs,steininger,dulny,hotho}@informatik.uni-wuerzburg.de

## Abstract

*Recent studies have shown that many deep metric learning loss functions perform very similarly under the same experimental conditions. One potential reason for this unexpected result is that all losses let the network focus on similar image regions or properties. In this paper, we investigate this by conducting a two-step analysis to extract and compare the learned visual features of the same model architecture trained with different loss functions: First, we compare the learned features on the pixel level by correlating saliency maps of the same input images. Second, we compare the clustering of embeddings for several image properties, e.g. object color or illumination. To provide independent control over these properties, photo-realistic 3D car renders similar to images in the Cars196 dataset are generated. In our analysis, we compare 14 pretrained models from a recent study and find that, even though all models perform similarly, different loss functions can guide the model to learn different features. We especially find differences between classification and ranking based losses. Our analysis also shows that some seemingly irrelevant properties can have significant influence on the resulting embedding. We encourage researchers from the deep metric learning community to use our methods to get insights into the features learned by their proposed methods.*

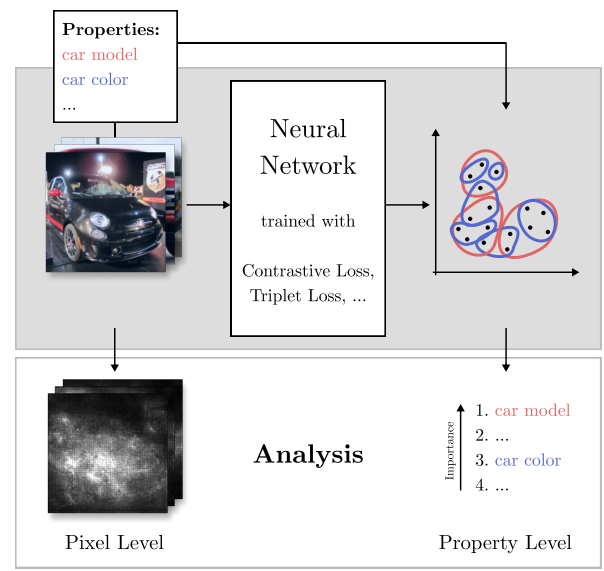

Figure 1. Given the standard DML setting with a neural network that maps input images to an embedding space (grey box), we propose two analysis methods. First, we identify pixels that are important for the network to create the embedding. We then compare them qualitatively and quantitatively between loss functions. Second, we investigate the influence of image properties on the clustering behavior in the embedding space and compare them between loss functions.

## 1. Introduction

In deep metric learning (DML), a neural network is trained to map input images to *m*-dimensional embedding vectors, that should be close to each other if the corresponding inputs share a given class. Thus, the network has to learn to extract discriminating input features to embed an image. Many loss functions have been introduced that can be categorized into ranking based losses [12], classification based losses [47], and hybrid methods combining both approaches [13]. Ranking based losses compare pairs, triplets, or higher-order tuples of data points to calculate a loss. Classification based methods usually learn one or multiple class representations and train the network to map inputs to the corresponding class embeddings.

In recent studies, different DML loss functions were shown to lead to similar test performances if compared fairly [23, 26]. Musgrave *et al.* [23] identify flaws in the evaluation settings of many DML papers and conduct a fair comparison between DML methods by testing several common loss functions with the same benchmark datasets, architecture, and test metrics. Their study finds very similar performances for all of the tested losses. In general, research has shown that even with similar performance, neural networks might learn to focus on different [5] and some-times even undesired input features [18] to form an output.

In this paper, we analyze and compare what features are paid attention to by neural networks trained with common DML loss functions. We propose two new analysis methods to get insights into the underlying processes of the network (see Figure 1) and apply them to the 14 pretrained models provided by Musgrave *et al*. (shown in Table 1).

The first proposed analysis method adapts a gradient-based explanation approach to DML, highlighting image pixels that lead the network to output the image representation [30]. Such visualizations can be used to make qualitative statements about the learned features on pixel level. Quantifying differences between loss functions is then possible by computing metrics inspired by the visual saliency literature [19, 25]. In our experiments, we identify a large difference between classification and ranking based loss functions on the Stanford Online Products dataset [33].

Our second proposed analysis method measures the influence of image properties, *e.g.* the rotation or color of an object, on the embeddings. Usually in DML, networks learn to differentiate one specific property, *e.g.* the car model for the Cars196 dataset [17], such that images of the same car model have similar embeddings and different car models are farther apart in embedding space. For testing, the network's ability to cluster new car images regarding their model is measured. Due to its training objective, other properties such as a car's color or environmental illumination should have minimal influence on the embedding, since the dataset contains images of the same car model in different colors and in different lighting conditions. If the network makes use of a property to output an embedding, images of the same property are likely to be clustered as well, potentially less pronounced. We propose to measure the clustering behavior of embeddings regarding different image properties to assess their importance. To ensure the properties are not correlated, we generate a large image dataset consisting of photo-realistic car renders. Since measuring the clustering behavior with the common metric R-Precision depends on the number of possible property values, we propose a property-independent extension, *Normalized R-Precision*, that enables the comparison of multiple properties at once. Our experiments show that properties used by the models are fairly consistent across all loss functions for our generated dataset. Surprisingly, some undesired properties show significant influence on the embedding.

Our contributions are: 1. We propose two methods to analyze the learned features of DML methods, one on pixel level and one on image property level. 2. We introduce a new measure called Normalized R-Precision making it possible to compare the influence of different image properties and a large dataset of 3D car renders with known properties. 3. Applying our new methods, we inspect 14 common DML loss functions and find that classification and ranking approaches tend to learn different features, depending on the dataset. 4. We make our code and data available to enable researchers to better understand their proposed methods.[1]

The paper is structured as follows: Section 2 discusses related work and the setup of our experiments. In Section 3, we describe and apply our pixel level analysis method. The property analysis is conducted in Section 4. Section 5 and Section 6 discuss and conclude our work.

## 2. Background

### 2.1. Deep Metric Learning

Deep metric learning (DML) aims to train a deep neural network to map input data onto an $m$-dimensional manifold, such that close representations mean high input data similarity. DML has been applied in many computer vision tasks such as image clustering, retrieval, person re-identification, and face verification, but also in other domains, *e.g.* 3D shape retrieval, semantic textual similarity, and speaker verification [15]. Mainly, there are three categories for DML loss functions: *Ranking* based methods rely on ranking pairs, triplets, or higher-order tuples of items [12, 43]. Since the selection of tuples is crucial for a stable training process, effective sampling strategies are also subject to research [44, 27, 10]. *Classification* based methods usually represent each class by one or multiple vectors that all class items should be mapped to. While this speeds up training without the need for specialized sampling strategies, representations are usually less detailed due to having only few representations per class [7, 47]. Consequently, *hybrid methods* combine ranking and classification approaches to train fast while capturing data details [13].

Changing evaluation settings between papers makes it difficult to fairly compare the effectiveness of different methods. Thus, recent work has started to compare DML loss functions under the same training and testing conditions, especially Musgrave *et al*. [23] whose pretrained models we use in this work. They find that reported improvements for newly presented methods are often too optimistic due to changes or flaws in the evaluation settings. On a leveled ground, the tested losses perform mostly similar.

### 2.2. Feature Analysis

**Pixel Level Features** Many works propose explanation methods for convolutional neural networks (CNNs) solving classification tasks [30, 9, 48, 49, 45, 31]. These methods highlight input pixels that have encouraged the model's decision for *one* class. DML networks learn multi-dimensional representations, making most of these methods non-trivial to apply. In DML, most work on feature importance focuses on the image retrieval task and aims to highlight areas that were the reason for the respective similarity score [35, 11, 50, 28]. Stylianou *et al.* [35] base their work on the *class activation maps* (CAM) [49] method, which calculates the dot product of each pixel in the last convolutional activation map with the other image's representation, resulting in a low resolution image showing what regions were deemed similar. This method is only applicable to CNNs that use global average pooling after the last convolution to create an embedding. CAM's extension Grad-CAM [28] takes the gradient of additional fully connected layers into account. Adapted to the DML setting by Zhu *et al.* [50], the activation map's pixels of two images are compared to each other, thus matching similar regions between image pairs. The method proposed by Chen *et al.* [3] uses triplets of training images and saves Grad-CAM's saliency maps to a database along with the corresponding embeddings. For test images, saliency maps of similar images from the database are interpolated.

---
[1]https://github.com/konstantinkobs/DML-analysis

While explaining similarities between image pairs is useful for tasks like image retrieval, we want to assess what image features are used by the model to embed one image. Our proposed method works with any neural network architecture and computes importances on pixel level.

**Property Level Features** Our method for analyzing the influence of image properties on the embedding uses a synthetic dataset. While prior work trained neural networks on synthetic datasets to improve performance in real-world scenarios [36, 16], only few works explored the use of synthetic data to analyze machine learning models trained on real-world data. Steininger *et al.* [34] generate fake OpenSteetMap images to assess the influence of properties such as street width or position on the output of a trained land use regression model. This approach is similar to our method, but we generate photo-realistic 3D images of cars and assess the clustering properties of the embeddings given a fixed property, in contrast to changing only one property.

### 2.3. Setup

For our analysis, we use trained models provided by Musgrave *et al.*, who train a BatchNorm Inception network [14] with 14 DML loss functions and compare their performance [23]. Table 1 lists all used losses. Each network outputs an 128 dimensional embedding per image and is trained and evaluated on three common datasets: **Cars196** [17] showing different car models (16 185 images/196 classes); **CUB200** [38] showing bird species (11 788 images/200 classes); and **Stanford Online Products (SOP)** [33] showing *ebay* products (120 053 images/22 634 classes). For our pixel level analysis, we use all three datasets, while for the image property analysis, we only analyze models trained on the Cars196 dataset, since we loosely imitate this dataset using generated car images.

For each loss, four trained models are provided, one for each fold of a four-fold cross validation performed to optimize hyperparameters using a Bayesian optimizer. All models are trained under the same conditions and the test results for all folds are averaged. We also report average results, since for all folds, the results are very similar. More information about the training setup and best hyperparameters of our used models can be found in [23]. In addition, we also add an untrained model "None", which is initialized using weights from an ImageNet [6] classifier, while the last layer is initialized with random weights [23].

| Method | Year | Loss type | Distance/Similarity |
|---|---|---|---|
| Contrastive [12] | 2006 | Ranking | Euclidean Distance |
| Triplet [43] | 2006 | Ranking | Euclidean Distance |
| NTXent [32] | 2016 | Ranking | Cosine Similarity |
| ProxyNCA [22] | 2017 | Classification | Squared Euclidean |
| Margin [44] | 2017 | Ranking | Euclidean Distance |
| Margin / class [44] | 2017 | Ranking | Euclidean Distance |
| Normalized Softmax [20, 40, 47] | 2017 | Classification | Dot Product Similarity |
| CosFace [39, 41] | 2018 | Classification | Cosine Similarity |
| ArcFace [7] | 2019 | Classification | Cosine Similarity |
| FastAP [2] | 2019 | Ranking | Squared Euclidean |
| SNR Contrastive [46] | 2019 | Ranking | SNR Distance |
| Multi Similarity [42] | 2019 | Ranking | Cosine Similarity |
| Multi Similarity + Miner [42] | 2019 | Ranking | Cosine Similarity |
| SoftTriple [24] | 2019 | Classification | Cosine Similarity |
| None | | | |

Table 1. Analyzed losses taken from [23]. Method "None" has the same architecture but without any training, thus only using weights from the feature extractor pre-trained on ImageNet.

### 2.4. Notation

For a given loss function $\ell \in \{\ell_{\text{contrastive}}, \ell_{\text{triplet}}, \dots\}$, a neural network $f_\ell : I \to \mathbb{R}^m$ maps an input image $\mathbf{I}$ from the dataset $I = \{\mathbf{I}_1, \dots, \mathbf{I}_n\}$ to the $m$-dimensional embedding space. This results in the embeddings $X = \{\mathbf{x}_1, \dots, \mathbf{x}_n\}$ with $\mathbf{x}_i = f_\ell(\mathbf{I}_i)$ for $i \in \{1, \dots, n\}$. Each image has properties; property $k$ has the possible values $A_k$. The value of image $\mathbf{I}_i$'s property $k$ is $a_k(\mathbf{I}_i) \in A_k$. One property $a_{\text{class}}(\mathbf{I}_i) \in A_{\text{class}}$ is the class of the image $\mathbf{I}_i$ that is used to define if two images are similar to each other while training (if $a_{\text{class}}(\mathbf{I}_i) = a_{\text{class}}(\mathbf{I}_j)$). A loss-specific distance function $d_\ell : X \times X \to \mathbb{R}_0^+$ calculates the distance for two embeddings, *e.g.* the Euclidean distance. While $d_\ell$ can also be a similarity measurement, *e.g.* cosine similarity, that should be maximized between similar embeddings, we assume it to be a distance metric in the rest of this paper for brevity.

## 3. Analysis on Pixel Level

### 3.1. Saliency Maps

Our first proposed method aims to identify features on pixel level that are important for the network's decision to output a certain embedding. For this, we adapt a gradient based explanation method to the DML setting to derive so-called saliency maps [30]. They are used to qualitatively

10646

|  |  | Ranking |  |  |  |  |  |  |  |  | Classification |  |  |  |  |  |
|---|---|---|---|---|---|---|---|---|---|---|---|---|---|---|---|---|---|
|  |  | Contrastive | Triplet | NTXent | Margin | Margin/class | FastAP | SNR Con. | MS | MS+Miner | ProxyNCA | N. Softmax | CosFace | ArcFace | SoftTriple | None |
| Ranking | Contrastive |  | 62±12 | 59±12 | 61±12 | 61±12 | 59±13 | 60±12 | 59±13 | 60±13 | 46±20 | 49±17 | 45±20 | 45±20 | 46±20 | 51±21 |
|  | Triplet | 62±12 |  | 63±12 | 66±11 | 66±12 | 63±13 | 61±13 | 63±12 | 64±12 | 50±19 | 52±18 | 50±20 | 49±20 | 50±20 | 59±20 |
|  | NTXent | 59±12 | 63±12 |  | 61±12 | 62±12 | 59±13 | 57±14 | 59±13 | 62±12 | 52±18 | 54±16 | 52±18 | 52±18 | 53±17 | 56±19 |
|  | Margin | 61±12 | 66±11 | 61±12 |  | 65±12 | 61±13 | 59±13 | 62±12 | 62±12 | 50±19 | 51±17 | 50±19 | 49±19 | 50±19 | 58±19 |
|  | Margin/class | 61±12 | 66±12 | 62±12 | 65±12 |  | 61±13 | 59±13 | 62±13 | 63±13 | 50±19 | 52±17 | 50±20 | 49±20 | 50±20 | 58±20 |
|  | FastAP | 59±13 | 63±13 | 59±13 | 61±13 | 61±13 |  | 61±13 | 61±13 | 60±13 | 49±20 | 52±17 | 50±20 | 49±20 | 50±19 | 55±21 |
|  | SNR Con. | 60±12 | 61±13 | 57±14 | 59±13 | 59±13 | 61±13 |  | 59±14 | 57±14 | 44±22 | 47±19 | 43±22 | 43±22 | 44±21 | 48±23 |
|  | MS | 59±13 | 63±12 | 59±13 | 62±12 | 62±13 | 61±13 | 59±14 |  | 62±13 | 53±18 | 54±16 | 53±18 | 53±18 | 53±18 | 58±18 |
|  | MS+Miner | 60±13 | 64±12 | 62±12 | 62±12 | 63±13 | 60±13 | 57±14 | 62±13 |  | 54±17 | 55±16 | 54±18 | 53±18 | 54±17 | 58±18 |
| Classif. | ProxyNCA | 46±20 | 50±19 | 52±18 | 50±19 | 50±19 | 49±20 | 44±22 | 53±18 | 54±17 |  | 62±14 | 67±12 | 66±12 | 64±13 | 60±16 |
|  | N. Softmax | 49±17 | 52±18 | 54±16 | 51±17 | 52±17 | 52±17 | 47±19 | 54±16 | 55±16 | 62±14 |  | 64±13 | 63±14 | 63±13 | 59±16 |
|  | CosFace | 45±20 | 50±20 | 52±18 | 50±19 | 50±20 | 50±20 | 43±22 | 53±18 | 54±18 | 67±12 | 64±13 |  | 69±11 | 67±11 | 63±14 |
|  | ArcFace | 45±20 | 49±20 | 52±18 | 49±19 | 49±20 | 49±20 | 43±22 | 53±18 | 53±18 | 66±12 | 63±14 | 69±11 |  | 65±12 | 61±15 |
|  | SoftTriple | 46±20 | 50±20 | 53±17 | 50±19 | 50±20 | 50±19 | 44±21 | 53±18 | 54±17 | 64±13 | 63±13 | 67±11 | 65±12 |  | 62±15 |
|  | None | 51±21 | 59±20 | 56±19 | 58±19 | 58±20 | 55±21 | 48±23 | 58±18 | 58±18 | 60±16 | 59±16 | 63±14 | 61±15 | 62±15 |  |

Table 2. Correlations between all loss functions on the SOP dataset. All values are given in percent. Larger values have darker cells.

|  | Ranking | Classification | None |
|---|---|---|---|
| **Ranking** | 86±6 | 85±6 | 75±11 |
| **Classification** | 85±6 | 86±5 | 74±11 |
| **None** | 75±11 | 74±11 | 100±0 |

Table 3. Correlations between all loss functions on the Cars196 dataset. All values are given in percent.

|  | Ranking | Classification | None |
|---|---|---|---|
| **Ranking** | 90±6 | 90±6 | 87±6 |
| **Classification** | 90±6 | 90±6 | 87±6 |
| **None** | 87±6 | 87±6 | 100±0 |

Table 4. Correlations between all loss functions on the CUB200 dataset. All values are given in percent.

analyze one network on a single image, however, we propose quantitative measures to compare models.

Our saliency maps seek to answer the question "What are the main image regions that guided the network to output the specific embedding?". Intuitively, we obtain the final embedding $\mathbf{x}_i$ by altering the pixels of an image that shows no features, i.e. a completely black image $\mathbf{I}_{\text{base}}$, towards $\mathbf{I}_i$. The larger the change towards the final embedding, the more important a pixel. Thus, we want to identify the most influential pixels regarding the distance $d_\ell(\mathbf{x}_i, \mathbf{x}_{\text{base}})$ between the image's embedding $\mathbf{x}_i$ and the embedding of the black image $\mathbf{x}_{\text{base}} = f_\ell(\mathbf{I}_{\text{base}})$. We do this by computing the gradients of the loss-specific distance w.r.t. the input $\mathbf{I}_i$:

$$\mathbf{s}_\ell(\mathbf{I}_i) = \partial d_\ell(\mathbf{x}_i, \mathbf{x}_{\text{base}}) / \partial \mathbf{I}_i. \quad (1)$$

Since these gradients can be noisy, we apply the Smooth-Grad method [31] by creating $l$ image variants by adding gaussian noise $\mathcal{N}(0, \sigma^2)$ to the input image and averaging the resulting gradients:

$$\hat{\mathbf{s}}_\ell(\mathbf{I}_i) = \frac{1}{l} \sum_1^l \mathbf{s}_\ell\left(\mathbf{I}_i + \mathcal{N}(0, \sigma^2)\right). \quad (2)$$

High absolute gradients indicate that changing the corresponding input value has large influence on the measured distance, thus identifying pixels responsible for the deviation of the base embedding. We post-process the gradients using common techniques, namely (in this order) taking the absolute value, averaging across the color channel dimension, clipping values higher than the 99[th] percentile, and scaling the values to a range from zero to one. These steps make the raw gradients more semantic, yielding an interpretable saliency map $\tilde{\mathbf{s}}_\ell(\mathbf{I}_i)$ [31].

Overall, this method is a qualitative technique to highlight important image areas for the network. While this can be used to visualize differences between DML loss functions on single images, we propose to quantify differences using this technique: Given two models $f_{\ell_1}$ and $f_{\ell_2}$ trained with different losses $\ell_1$ and $\ell_2$, we apply both models on the same test images $\mathbf{I}_1, \ldots, \mathbf{I}_n$ and compute the saliency maps $\tilde{\mathbf{s}}_\ell(\mathbf{I}_1), \ldots, \tilde{\mathbf{s}}_\ell(\mathbf{I}_n)$ for $\ell_1$ and $\ell_2$. Inspired by the literature for the visual saliency task [19, 25], i.e. estimating a heatmap of a human's eye fixations on an image, we compare saliency maps by calculating the average Pearson product-moment correlation coefficient and Jensen-Shannon Divergence (JSD) [8] between the same image's saliency maps of two networks. We transform correlations to Fisher-Z space before averaging [29] and divide each saliency map by its sum to obtain probability distributions for JSD. Mean correlations close to one show that both saliency maps usually have a linear dependency, meaning that both networks attend to the same image regions. Lower values indicate that both models learned different features in order to represent images. A mean JSD of zero means that both methods produce the same saliency maps, while higher values (bounded by 1, due to base 2 logarithm) show larger differences.

### 3.2. Experiments

**Quantitative Results** Tables 2 to 4 show the correlations' means and standard deviations for the test datasets of SOP,

Cars196, and CUB200, respectively. We omit the Jensen-Shannon Divergence and their standard deviations in all tables for legibility, as all values are around 0.02 ± 0.01 and show similar tendencies as the correlation. We also show condensed tables for Cars196 and CUB200 for brevity. The full tables can be found in Appendix A.

Compared to the other datasets, correlations for SOP are generally weaker with larger standard deviations, showing that losses are not consistent across images in terms of feature extraction. Surprisingly, loss pairs of different loss types (ranking *vs.* classification) show lower correlations than pairs of the same loss type, suggesting that different loss types lead to different saliency maps. Grouping loss pairs by their distance/similarity metric does not show such clear differences. The "None" model has stronger correlation with classification than with ranking losses, which is expected due to its training on a classification task.

For the Cars196 and CUB200 datasets, we average the table entries for each loss type combination, since these show almost the same correlations with each other. The strong correlations show that models tend to focus on the same pixels to embed an image. Also, the standard deviation is around 0.06 for both datasets, indicating consistent behavior across all images. We can not identify the same large drop in correlation when comparing ranking and classification losses. Only a small tendency is present for the Cars196 dataset. A noticeable drop in correlation can only be observed with the untrained "None" model, which is expected, but surprisingly not that steep. The strong correlations between trained models and the untrained "None" model show high similarity in extracted features between them. We conclude that ImageNet based initialization of the untrained models already leads to features that are picked up by the analyzed DML networks.

**Qualitative Results** Given the findings of the quantitative results, we now visually inspect the learned features. Figure 2 shows saliency maps for all investigated networks for a sample image from the SOP test dataset, showing a chair. We observe that most methods highlight parts of the chair, but focus on different areas. While *e.g.* Contrastive Loss attends to the chair's legs and back, CosFace pays more attention to the seat. Given the quantitative difference between ranking and classification losses, we observe that ranking based methods usually show more pronounced local highlights, while classification based methods highlight broader areas. It also seems that the background area is more important for classification approaches. This observation is difficult to verify since we do not have any segmentation information about the image's fore- and background. Also, while humans interpret saliency maps in terms of concepts like foreground, background, or object parts, the neural network only works on pixel level. To be able to make statements about the influence of such properties, we continue with the second step in our analysis.

## 4. Analysis on Property Level

### 4.1. Property Clusters

Image properties describe concepts like object form, color, or orientation. We investigate the question "What image properties influence the model output?". Each input image has a set of properties and their values. A property with high influence on the embedding fulfills the two clustering objectives: First, fixing this property and changing all other property values should result in small deviations in embedding space. Second, changing the property while keeping everything else fixed should result in large deviations in embedding space. This idea is used in common evaluation metrics in DML, but are only applied to the image's "class" property. A DML neural network is considered to work well, if it maps test images with the same class to similar locations in embedding space, while embedding images with different classes to different locations. For Cars196, a test class is a certain *car model*, for which many different images from different angles, car colors, *etc.* exist in the dataset. Instead of the "class", we use other image properties such as the car's color or the car's orientation. Even though the neural network has not been trained on these types of data splits, we can still measure the closeness of the resulting embeddings regarding the defined property. If, for example, grouping embeddings by car orientation shows well-defined clusters, we can conclude that changing the orientation has significant effect on the network's output. If the network is invariant to the car's orientation, changing it does not significantly alter the embedding vector, thus showing no clustering behavior. Clustering examples for three properties are shown in Figure 3.

In order to measure the clustering behavior of properties, we propose to use the common DML metric R-Precision as the base. For one query embedding $\mathbf{x}_q$ and a property $k$, the $R_{k,q}$ closest embeddings from the dataset are retrieved, where $R_{k,q}$ is the number of images with the same property value $a_k(\mathbf{x}_i)$ in the dataset:

$$\mathcal{F}^R_{\mathbf{x}_q} = \operatorname*{arg\,min}_{\mathcal{F} \subset X, |\mathcal{F}| = R_{k,q}} \sum_{\mathbf{x}_i \in \mathcal{F}} d_\ell(\mathbf{x}_q, \mathbf{x}_i) \,. \quad (3)$$

The R-Precision (R-Prec) is then defined as

$$\text{R-Prec}_k = \frac{1}{n} \sum_{q=1}^n \frac{\left|\{\mathbf{x}_i \in \mathcal{F}^R_{\mathbf{x}_q} \mid a_k(\mathbf{I}_i) = a_k(\mathbf{I}_q)\}\right|}{R_{k,q}} \,, \quad (4)$$

*i.e.* the average fraction of items having the same property value. This metric measures how well the model puts items with the same property value closer together. The higher

10648

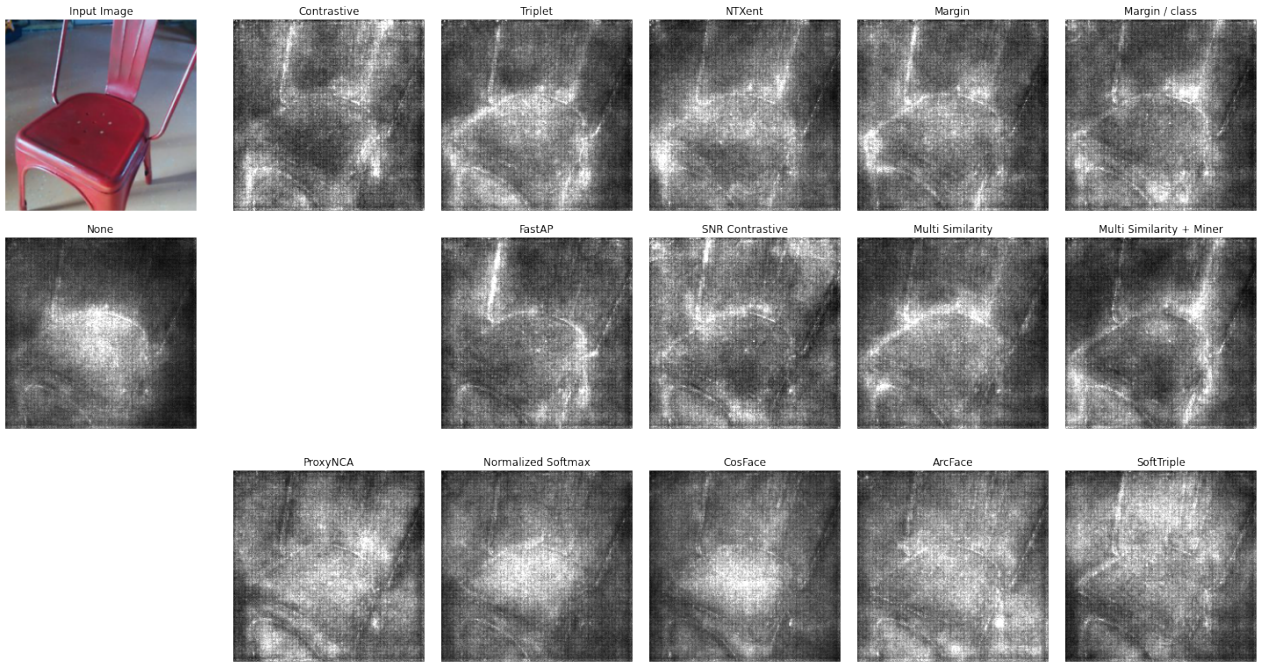

Figure 2. Saliency maps of a sample image from SOP. The original image and the "None" baseline (pre-trained ImageNet weights) are in the first column. The first two rows show *embedding* losses, the third row shows *classification* losses. More samples are in Appendix B.

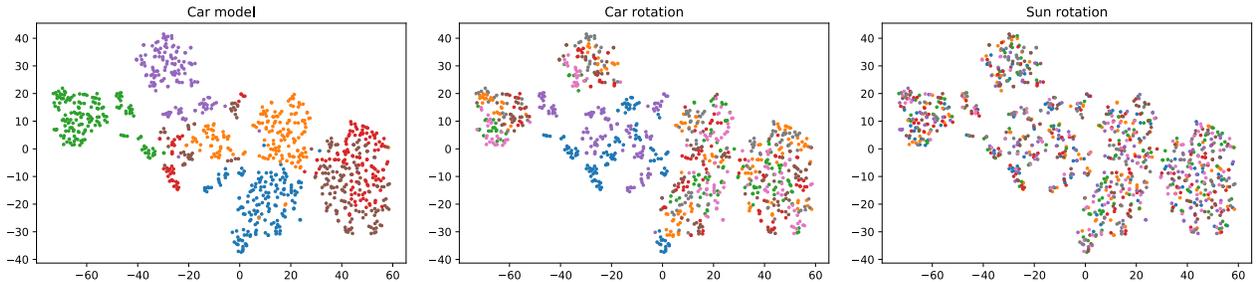

Figure 3. 1 000 embeddings from a Cars196 model with Contrastive Loss visualized with t-SNE [37]. Color denotes different property values. *Car model* is clustered well which shows that the model uses this property as a discriminating feature for its embedding output. *Car rotation* shows local clusters, thus still having an influence on the embedding. *Sun rotation* has no influence and is not clustered at all.

the R-Precision for a certain property, the better the embedding clusters w.r.t. to this property. Altering the property thus significantly changes the embedding vectors, while the network is less influenced by other properties.

However, R-Precision depends on the number of property values: Given a random embedding and only two possible property values with the same number of items, the expected R-Precision is 0.5. For a property with ten possible values, the expected R-Prec score is 0.1. Thus, an absolute comparison between properties is not possible, as random embeddings would score differently. Therefore, we propose to apply a normalization step to the R-Precision calculation. With randomly generated embeddings for all $n$ images, the number of images with the same property value of the property $k$ as the query embedding $\mathbf{x}_q$ is binomially distributed. For the metric calculation, we take $R_{k,q}$ samples. There is a probability of $p_{k,q} = \frac{|\{\mathbf{x}_i \in X | a_k(\mathbf{I}_i) = a_k(\mathbf{I}_q)\}|}{n}$ that a close embedding has the same property value. We use the mean $\mu_{k,q} = R_{k,q} \cdot p_{k,q}$ and standard deviation $\sigma_{k,q} = R_{k,q} \cdot p_{k,q} \cdot (1 - p_{k,q})$ of this binomial distribution to normalize the R-Precision calculation per query embedding. We obtain the *Normalized R-Precision (NR-Prec)*:

$$\text{NR-Prec}_k = \frac{1}{n} \sum_{q=1}^{n} \frac{\left|\{\mathbf{x}_i \in \mathcal{F}_{\mathbf{x}_q}^R \mid a_k(\mathbf{I}_i) = a_k(\mathbf{I}_q)\}\right| - \mu_{k,q}}{\sigma_{k,q}}.$$
(5)

NR-Prec is zero if the clustering is as good as for random embeddings. The larger the deviation from zero, the lower the probability of this clustering being due to randomness. Due to the normalization, we gain two advantages over R-Precision: On the one hand, we can now compare properties with different numbers of possible values, allowing us to sort different properties by how well the model picks them up. NR-Prec results and their ranking can also be compared between models in order to check if different loss functions pick up similar high-level features. On the other hand, it is possible to measure statistical significance. Given a sufficiently large dataset, the normalized binomial distribution approximates a normal distribution, so if NR-Prec exceeds 2.576, the embeddings locations are significantly different from random embeddings with a 1 % significance level.

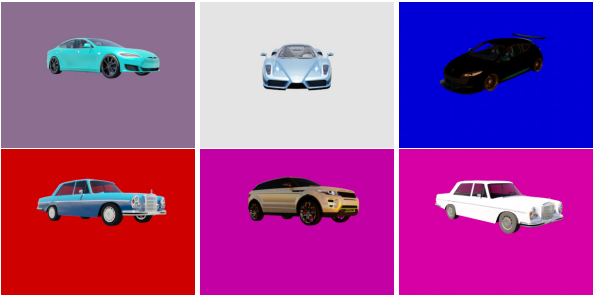

Figure 4. Sample images of our generated car dataset. We vary eleven properties, such as the model, lighting, and colors.

| Property | Possible Values |
|---|---|
| Car model | Ferrari Enzo, Mercedes Benz 300sel, Megane RS, Mercedes AMG Coupe, Range Rover Evoque, Tesla Model S |
| Car rotation | 0°, 45°, ..., 315° |
| Car color | |
|   Hue | 0.0, 0.1, ..., 0.9 |
|   Saturation | 0.0, 0.25, ..., 2.0 |
|   Value | 0.0, 0.25, ..., 2.0 |
| Background color | |
|   Hue | 0.0, 0.1, ..., 0.9 |
|   Saturation | 0.0, 0.25, ..., 2.0 |
|   Value | 0.0, 0.25, ..., 2.0 |
| Camera height | 0.5, 1.5, 2.5, 3.5 |
| Sun elevation | 0°, 45°, 90° |
| Sun rotation | 0°, 45°, ..., 315° |

Table 5. All properties and corresponding possible values in the car renderings. Combinations were chosen uniform at random.

It is important that properties in the dataset used to calculate NR-Prec are chosen independently for each image to correctly measure the property importance. Imagine that the dataset consists of car images where each car model has its own specific color. If the network is trained to cluster the car model, the high correlation between car model and color leads to well-clustered embeddings for the color, even though the network might only attend to car model features. In order to make sure that properties are independent of each other for each image in our test dataset, we create photo-realistic 3D renders of cars, loosely imitating the Cars196 dataset. Besides the *car model*, we alter properties such as the *car's color*, *rotation*, and *illumination*. Table 5 shows all altered properties with their possible values. We sample 100 000 from all possible combinations uniformly at random to ensure all splits have similar sizes and independent property value choices. Figure 4 shows samples of the dataset that we generated using Blender [4].

### 4.2. Experiments

The images of our generated dataset are fed through all tested models and the NR-Prec is computed for each property, giving the results in Table 6. All losses attend to the properties in the same order. The car's *model* yields the most notable embedding clusters, since all networks are trained to differentiate between car models. The *rotation*'s clusters likely stem from discriminating features being visible to the camera. For presumably similar reasons, the *camera height* shows good clustering as well. This might also be because only few Cars196 training images show the car from low perspectives. The *sun rotation* shows expectedly bad clustering behavior: Cars illuminated from many possible directions are seen during training. Surprisingly, the *sun's elevation* has high influence on the embedding. This might come from the training dataset consisting mostly of photos taken in daylight. A low *sun elevation* makes the light warmer and casts longer shadows. These influences on the image might be picked up by the DML models, since there are few training examples in this situation. The *background color* has negligible effect on the clustering, which is expected, since cars are pictured in many different environments. In contrast, the *car color* leads to embeddings significantly different from random embeddings, which is somewhat surprising, as each training *car model* is shown in multiple colors. We suspect that different colors make it more difficult to identify certain features.

We observe that the "None" baseline has the same property order as trained models. Compared to trained embeddings, the *car model* shows weaker clusters and the *background color* properties yield embeddings significantly different from random assignments. The network's weights, except for the last layer, are initialized with trained weights from the ImageNet classification task. The network's embedding therefore represents features that were important for image classification. For this task, the learned features are usually invariant to lighting conditions, but the environment can be a discriminating feature, *e.g.* the presence of water helps to identify ships [18]. Thus, the "None" network attends to the *background* properties more to generate embeddings. During fine-tuning on the Cars196 dataset, all loss functions guide the network to learn that the *background* is less important for embedding the *car model*.

|  |  | Car | | Car Color | | | Background Color | | | | Sun | |
|---|---|---|---|---|---|---|---|---|---|---|---|---|
|  |  | Model | Rotation | Hue | Saturation | Value | Hue | Saturation | Value | Camera Height | Elevation | Rotation |
| Ranking | Contrastive | 58.32 | 39.60 | 3.02 | 3.28 | 4.72 | 1.52 | 1.30 | 2.09 | 20.92 | 8.23 | 0.85 |
|  | Triplet | 57.36 | 37.37 | 3.25 | 3.26 | 4.69 | 1.63 | 1.43 | 2.50 | 19.50 | 8.22 | 0.74 |
|  | NTXent | 57.87 | 38.22 | 3.44 | 3.30 | 4.54 | 1.57 | 1.46 | 2.18 | 20.32 | 8.55 | 0.78 |
|  | Margin | 57.91 | 38.58 | 3.50 | 3.18 | 4.76 | 1.59 | 1.29 | 2.04 | 21.21 | 8.30 | 0.78 |
|  | Margin / class | 58.92 | 39.65 | 3.41 | 3.36 | 4.83 | 1.89 | 1.72 | 2.25 | 21.55 | 8.82 | 0.76 |
|  | FastAP | 55.84 | 38.44 | 2.81 | 3.10 | 4.87 | 1.05 | 1.24 | 1.81 | 20.49 | 7.45 | 0.71 |
|  | SNR Contrastive | 57.38 | 39.88 | 3.41 | 3.37 | 5.05 | 1.69 | 1.56 | 2.22 | 21.40 | 8.66 | 0.82 |
|  | Multi Similarity | 59.81 | 41.03 | 3.18 | 3.37 | 4.95 | 1.83 | 1.74 | 2.59 | 21.14 | 8.73 | 0.87 |
|  | Multi Similarity + Miner | 57.82 | 38.86 | 3.24 | 3.13 | 4.52 | 1.84 | 1.52 | 2.03 | 20.03 | 7.66 | 0.77 |
| Classif. | ProxyNCA | 57.68 | 37.64 | 4.73 | 3.93 | 5.72 | 2.33 | 2.09 | 2.52 | 19.82 | 9.77 | 0.84 |
|  | Normalized Softmax | 57.26 | 37.76 | 3.85 | 3.97 | 5.44 | 1.68 | 1.80 | 2.31 | 19.43 | 8.81 | 0.80 |
|  | CosFace | 56.50 | 38.51 | 4.00 | 3.65 | 5.32 | 2.40 | 2.39 | 2.64 | 19.10 | 8.72 | 0.83 |
|  | ArcFace | 55.62 | 37.20 | 4.15 | 3.72 | 4.91 | 2.90 | 2.84 | 2.93 | 18.70 | 8.98 | 0.78 |
|  | SoftTriple | 57.36 | 38.08 | 3.63 | 3.81 | 5.37 | 1.81 | 2.00 | 2.25 | 19.72 | 8.43 | 0.77 |
|  | None | 50.61 | 36.25 | 4.32 | 3.92 | 4.48 | 4.81 | 4.09 | 6.80 | 20.75 | 8.64 | 0.75 |
|  | Ranking Mean | 57.91 | **39.07** | 3.25 | 3.26 | 4.77 | 1.62 | 1.48 | 2.19 | **20.73** | 8.29 | 0.79 |
|  | Classification Mean | 56.89 | 37.84 | **4.07** | **3.82** | **5.35** | **2.23** | **2.22** | **2.53** | 19.35 | **8.94** | 0.80 |

Table 6. NR-Precisions for the rendered car images. The higher the value (darker the cell shade), the less likely that the performance stems from randomly sampling neighbors. Significantly different values are underlined. We also give means for ranking and classification losses. There, bold text indicates that, on average, one loss type pays significantly more attention to this property than the other loss type.

When grouped by their loss type, we identify differences in NR-Prec scores between ranking *vs*. classification. We apply a Mann-Whitney U test [21] with a significance level of 1 %, showing significant differences between classification and ranking based loss functions for all image properties except the *car model* and the *sun rotation*. While ranking loss functions show significantly larger influence of *car rotation* and *camera height*, classification based loss functions attend to the *car and background color* properties as well as the *sun elevation* significantly more than ranking losses. This supports our qualitative observation that, on average, the *background* properties tend to play a larger role in classification than in ranking losses.

## 5. Discussion

In general, our experiments have shown that all networks learn similar features on pixel as well as property level. On the SOP dataset, however, classification based losses attend to fairly different regions than ranking based losses. Based on the DML objective to cluster images from the same class together, we would expect models to be invariant to unimportant features for the class, *e.g.* the car's color, its orientation, or environmental illumination when trained to embed the car model. However, we have shown that the properties *car color*, *rotation*, *sun's elevation*, and *camera height* have significant influence on the embeddings. Also, classification losses usually pay more attention to the background of images than ranking losses. Our proposed methods serve as tools to analyze what features are learned by DML neural networks and to evaluate if they are invariant to unimportant properties. Our tools can be used to develop and evaluate methods that encourage invariance for undesired properties, *e.g.* [1]. Simple pre-processing steps, like hue shifts, grayscaling, or skewing, could remove dependencies on the car's color or camera heights/angles. While we have not investigated the influence of other camera parameters such as focal length or image properties such as contrast or brightness, these might also have undesired effects on the resulting embeddings. Correcting for such parameters methodologically is desirable.

Since we found differences between classification and ranking based methods, future work might analyze hybrid loss functions and find reasons for found differences. Besides losses, our methods are able to examine differences between other methodological choices, *e.g.* tuple mining or regularization methods. Our saliency map based approach is dataset agnostic and can be applied to any trained DML model and input image. Since Cars196 is a common DML benchmark, most researchers already train models on this dataset. Our image property analysis can thus be conducted without additional training. We encourage researchers to use our tools to gain insight into their proposed methods.

## 6. Conclusion

In this paper we have analyzed 14 different deep metric learning losses regarding their learned features on pixel as well as image property level. For this, we have proposed two methods, one based on saliency maps highlighting pixels responsible for the image embedding and one based on the clustering behavior of image properties. We were able to show that ranking based and classification based losses guide the network to learn different features, depending on the dataset. We also have found that all losses pay attention to seemingly undesired properties such as the car's color or the sun elevation. Our two proposed methods are the base for further comparisons of deep metric learning methods.

# Appendix for "Do Different Deep Metric Learning Losses Lead to Similar Learned Features?"

Konstantin Kobs    Michael Steininger    Andrzej Dulny    Andreas Hotho
University of Würzburg
Germany
{kobs,steininger,dulny,hotho}@informatik.uni-wuerzburg.de

## A. Full Results Tables of the Quantitative Pixel Level Analysis

Tables 1 and 2 show the full correlation tables for our quantitative analysis. Tables 3 to 5 show the Jensen-Shannon Divergence tables.

|  |  | Ranking |  |  |  |  |  |  |  |  | Classification |  |  |  |  |  |
|---|---|---|---|---|---|---|---|---|---|---|---|---|---|---|---|---|
|  |  | Contrastive | Triplet | NTXent | Margin | Margin/class | FastAP | SNR Con. | MS | MS+Miner | ProxyNCA | N. Softmax | CosFace | ArcFace | SoftTriple | None |
| Ranking | Contrastive |  | 84±6 | 86±5 | 86±5 | 85±5 | 85±5 | 86±5 | 86±5 | 87±5 | 85±5 | 86±5 | 85±5 | 85±5 | 86±5 | 74±11 |
|  | Triplet | 84±6 |  | 85±6 | 84±6 | 85±6 | 84±6 | 84±6 | 84±6 | 84±6 | 83±6 | 84±6 | 82±7 | 82±7 | 84±6 | 75±10 |
|  | NTXent | 86±5 | 85±6 |  | 85±6 | 85±5 | 85±6 | 85±5 | 86±5 | 86±5 | 85±5 | 86±5 | 84±6 | 84±6 | 85±5 | 74±10 |
|  | Margin | 86±5 | 84±6 | 85±6 |  | 85±6 | 85±6 | 85±6 | 85±6 | 85±6 | 84±6 | 85±6 | 84±6 | 83±6 | 84±6 | 74±10 |
|  | Margin/class | 85±5 | 85±6 | 85±5 | 85±6 |  | 84±6 | 84±6 | 85±5 | 85±5 | 84±5 | 85±5 | 84±6 | 83±6 | 84±6 | 75±10 |
|  | FastAP | 85±5 | 84±6 | 85±6 | 85±6 | 84±6 |  | 85±6 | 85±5 | 85±5 | 85±5 | 85±5 | 84±6 | 84±6 | 85±5 | 73±11 |
|  | SNR Con. | 86±5 | 84±6 | 85±5 | 85±6 | 85±6 | 85±6 |  | 86±5 | 85±5 | 84±5 | 86±5 | 84±6 | 84±5 | 85±5 | 74±10 |
|  | MS | 86±5 | 84±6 | 86±5 | 85±5 | 85±5 | 85±5 | 86±5 |  | 86±5 | 85±5 | 86±5 | 85±6 | 85±5 | 86±5 | 74±10 |
|  | MS+Miner | 87±5 | 84±6 | 86±5 | 85±6 | 85±5 | 85±5 | 85±5 | 86±5 |  | 85±5 | 86±5 | 85±6 | 85±5 | 86±5 | 73±11 |
| Classif. | ProxyNCA | 85±5 | 83±6 | 85±5 | 84±6 | 84±5 | 85±5 | 84±5 | 85±5 | 85±5 |  | 87±4 | 85±5 | 85±5 | 86±5 | 74±11 |
|  | N. Softmax | 86±5 | 84±6 | 86±5 | 85±6 | 85±5 | 85±5 | 86±5 | 86±5 | 86±5 | 87±4 |  | 86±5 | 86±5 | 87±4 | 74±11 |
|  | CosFace | 85±5 | 82±7 | 84±6 | 84±6 | 84±6 | 84±6 | 84±6 | 85±6 | 85±6 | 85±5 | 86±5 |  | 85±5 | 87±4 | 73±11 |
|  | ArcFace | 85±5 | 82±7 | 84±6 | 83±6 | 83±6 | 84±6 | 84±5 | 85±5 | 85±5 | 85±5 | 86±5 | 85±5 |  | 86±4 | 73±11 |
|  | SoftTriple | 86±5 | 84±6 | 85±5 | 84±6 | 84±6 | 85±5 | 85±5 | 86±5 | 86±5 | 86±5 | 87±4 | 87±4 | 86±4 |  | 74±11 |
|  | None | 74±11 | 75±10 | 74±10 | 74±10 | 75±10 | 73±11 | 74±10 | 74±10 | 73±11 | 74±10 | 74±11 | 73±11 | 73±11 | 74±11 |  |

Table 1. Correlations between all loss functions on the Cars196 dataset. All values are in percent. Larger values have darker cells.

|  |  | Ranking |  |  |  |  |  |  |  |  | Classification |  |  |  |  |  |
|---|---|---|---|---|---|---|---|---|---|---|---|---|---|---|---|---|
|  |  | Contrastive | Triplet | NTXent | Margin | Margin/class | FastAP | SNR Con. | MS | MS+Miner | ProxyNCA | N. Softmax | CosFace | ArcFace | SoftTriple | None |
| Ranking | Contrastive |  | 88±7 | 89±6 | 89±6 | 89±5 | 89±6 | 89±5 | 89±5 | 89±5 | 88±6 | 89±6 | 89±6 | 89±6 | 89±5 | 86±6 |
|  | Triplet | 88±7 |  | 89±5 | 89±6 | 89±6 | 89±6 | 89±5 | 89±5 | 89±5 | 88±6 | 89±5 | 88±7 | 88±6 | 88±6 | 86±6 |
|  | NTXent | 89±6 | 89±5 |  | 89±5 | 90±5 | 90±5 | 89±5 | 90±5 | 90±4 | 89±5 | 90±5 | 88±6 | 89±5 | 89±5 | 86±6 |
|  | Margin | 89±6 | 89±6 | 89±5 |  | 90±5 | 90±5 | 90±5 | 90±5 | 90±5 | 89±5 | 89±5 | 88±6 | 89±6 | 89±6 | 86±6 |
|  | Margin/class | 89±5 | 89±6 | 90±5 | 90±5 |  | 90±5 | 90±5 | 90±5 | 90±5 | 89±6 | 89±5 | 88±6 | 89±5 | 89±5 | 86±6 |
|  | FastAP | 89±6 | 89±6 | 90±5 | 90±5 | 90±5 |  | 90±5 | 90±5 | 90±5 | 89±5 | 90±5 | 89±6 | 90±5 | 90±5 | 86±6 |
|  | SNR Con. | 89±5 | 89±5 | 89±5 | 90±5 | 90±5 | 90±5 |  | 90±4 | 90±5 | 89±5 | 90±5 | 89±5 | 89±5 | 89±5 | 87±6 |
|  | MS | 89±5 | 89±5 | 90±5 | 90±5 | 90±5 | 90±5 | 90±4 |  | 90±4 | 89±5 | 90±4 | 89±5 | 89±5 | 89±5 | 87±5 |
|  | MS+Miner | 89±5 | 89±5 | 90±4 | 90±5 | 90±5 | 90±5 | 90±5 | 90±4 |  | 89±5 | 90±5 | 89±5 | 89±5 | 89±5 | 86±6 |
| Classif. | ProxyNCA | 88±6 | 88±6 | 89±5 | 89±5 | 89±6 | 89±5 | 89±5 | 89±5 | 89±5 |  | 89±5 | 88±6 | 89±5 | 89±5 | 87±5 |
|  | N. Softmax | 89±6 | 89±5 | 90±5 | 89±5 | 89±5 | 90±5 | 90±5 | 90±4 | 90±5 | 89±5 |  | 89±6 | 90±5 | 90±5 | 86±6 |
|  | CosFace | 89±6 | 88±7 | 88±6 | 88±6 | 88±6 | 89±6 | 89±5 | 89±5 | 89±5 | 88±6 | 89±6 |  | 90±5 | 90±5 | 86±6 |
|  | ArcFace | 89±6 | 88±6 | 89±5 | 89±6 | 89±5 | 90±5 | 89±5 | 89±5 | 89±5 | 89±5 | 90±5 | 90±5 |  | 90±5 | 86±6 |
|  | SoftTriple | 89±5 | 88±6 | 89±5 | 89±6 | 89±5 | 90±5 | 89±5 | 89±5 | 89±5 | 89±5 | 90±5 | 90±5 | 90±5 |  | 86±6 |
|  | None | 86±6 | 86±6 | 86±6 | 86±6 | 86±6 | 86±6 | 87±6 | 87±5 | 86±6 | 87±5 | 86±6 | 86±6 | 86±6 | 86±6 |  |

Table 2. Correlations between all loss functions on the CUB200 dataset. All values are in percent. Larger values have darker cells.

|  | Ranking | | | | | | | | | Classification | | | | | |
|---|---|---|---|---|---|---|---|---|---|---|---|---|---|---|---|
|  | Contrastive | Triplet | NTXent | Margin | Margin/class | FastAP | SNR Con. | MS | MS+Miner | ProxyNCA | N. Softmax | CosFace | ArcFace | SoftTriple | None |
| Contrastive |  | 2±0 | 2±0 | 2±0 | 2±0 | 2±0 | 2±0 | 2±0 | 2±0 | 3±1 | 3±0 | 3±1 | 3±1 | 3±1 | 4±1 |
| Triplet | 2±0 |  | 2±0 | 2±0 | 2±0 | 2±0 | 2±0 | 2±0 | 2±0 | 3±1 | 3±0 | 3±1 | 3±1 | 3±1 | 3±1 |
| NTXent | 2±0 | 2±0 |  | 2±0 | 2±0 | 2±0 | 2±0 | 2±0 | 2±0 | 2±0 | 2±0 | 3±0 | 3±0 | 2±0 | 3±1 |
| Margin | 2±0 | 2±0 | 2±0 |  | 2±0 | 2±0 | 2±0 | 2±0 | 2±0 | 3±0 | 3±0 | 3±1 | 3±1 | 3±1 | 3±1 |
| Margin/class | 2±0 | 2±0 | 2±0 | 2±0 |  | 2±0 | 2±0 | 2±0 | 2±0 | 3±0 | 3±0 | 3±1 | 3±1 | 3±1 | 3±1 |
| FastAP | 2±0 | 2±0 | 2±0 | 2±0 | 2±0 |  | 2±0 | 2±0 | 2±0 | 3±1 | 3±0 | 3±1 | 3±1 | 3±1 | 3±1 |
| SNR Con. | 2±0 | 2±0 | 2±0 | 2±0 | 2±0 | 2±0 |  | 2±0 | 2±0 | 3±1 | 3±1 | 3±1 | 3±1 | 3±1 | 4±1 |
| MS | 2±0 | 2±0 | 2±0 | 2±0 | 2±0 | 2±0 | 2±0 |  | 2±0 | 3±0 | 3±0 | 3±0 | 3±0 | 3±0 | 3±1 |
| MS+Miner | 2±0 | 2±0 | 2±0 | 2±0 | 2±0 | 2±0 | 2±0 | 2±0 |  | 2±0 | 2±0 | 3±0 | 3±0 | 3±0 | 3±1 |
| ProxyNCA | 3±1 | 3±1 | 2±0 | 3±0 | 3±0 | 3±1 | 3±1 | 3±0 | 2±0 |  | 2±0 | 2±0 | 2±0 | 2±0 | 3±1 |
| N. Softmax | 3±0 | 3±0 | 2±0 | 3±0 | 3±0 | 3±0 | 3±1 | 3±0 | 2±0 | 2±0 |  | 2±0 | 2±0 | 2±0 | 3±1 |
| CosFace | 3±1 | 3±1 | 3±0 | 3±1 | 3±1 | 3±1 | 3±1 | 3±0 | 3±0 | 2±0 | 2±0 |  | 2±0 | 2±0 | 3±0 |
| ArcFace | 3±1 | 3±1 | 3±0 | 3±1 | 3±1 | 3±1 | 3±1 | 3±0 | 3±0 | 2±0 | 2±0 | 2±0 |  | 2±0 | 3±1 |
| SoftTriple | 3±1 | 3±1 | 2±0 | 3±1 | 3±1 | 3±1 | 3±1 | 3±0 | 3±0 | 2±0 | 2±0 | 2±0 | 2±0 |  | 3±0 |
| None | 4±1 | 3±1 | 3±1 | 3±1 | 3±1 | 3±1 | 4±1 | 3±1 | 3±1 | 3±1 | 3±1 | 3±0 | 3±1 | 3±0 |  |

Table 3. Jensen-Shannon Divergence between all loss functions on the SOP dataset. All values are in percent.

|  | Ranking | | | | | | | | | Classification | | | | | |
|---|---|---|---|---|---|---|---|---|---|---|---|---|---|---|---|
|  | Contrastive | Triplet | NTXent | Margin | Margin/class | FastAP | SNR Con. | MS | MS+Miner | ProxyNCA | N. Softmax | CosFace | ArcFace | SoftTriple | None |
| Contrastive |  | 2±0 | 1±0 | 2±0 | 2±0 | 2±0 | 2±0 | 1±0 | 1±0 | 2±0 | 2±0 | 2±0 | 2±0 | 2±0 | 3±1 |
| Triplet | 2±0 |  | 2±0 | 2±0 | 2±0 | 2±0 | 2±0 | 2±0 | 2±0 | 2±0 | 2±0 | 2±0 | 2±0 | 2±0 | 3±1 |
| NTXent | 1±0 | 2±0 |  | 2±0 | 1±0 | 2±0 | 1±0 | 1±0 | 2±0 | 2±0 | 1±0 | 2±0 | 2±0 | 2±0 | 3±1 |
| Margin | 2±0 | 2±0 | 2±0 |  | 2±0 | 2±0 | 2±0 | 2±0 | 2±0 | 2±0 | 2±0 | 2±0 | 2±0 | 2±0 | 3±1 |
| Margin/class | 2±0 | 2±0 | 1±0 | 2±0 |  | 2±0 | 2±0 | 2±0 | 2±0 | 2±0 | 2±0 | 2±0 | 2±0 | 2±0 | 3±1 |
| FastAP | 2±0 | 2±0 | 2±0 | 2±0 | 2±0 |  | 2±0 | 2±0 | 2±0 | 2±0 | 2±0 | 2±0 | 2±0 | 2±0 | 4±1 |
| SNR Con. | 2±0 | 2±0 | 1±0 | 2±0 | 2±0 | 2±0 |  | 1±0 | 2±0 | 2±0 | 1±0 | 2±0 | 2±0 | 2±0 | 3±1 |
| MS | 1±0 | 2±0 | 1±0 | 2±0 | 2±0 | 2±0 | 1±0 |  | 1±0 | 1±0 | 1±0 | 2±0 | 2±0 | 2±0 | 3±1 |
| MS+Miner | 1±0 | 2±0 | 2±0 | 2±0 | 2±0 | 2±0 | 2±0 | 1±0 |  | 2±0 | 1±0 | 2±0 | 2±0 | 1±0 | 3±1 |
| ProxyNCA | 2±0 | 2±0 | 2±0 | 2±0 | 2±0 | 2±0 | 2±0 | 1±0 | 2±0 |  | 1±0 | 1±0 | 2±0 | 1±0 | 3±1 |
| N. Softmax | 2±0 | 2±0 | 1±0 | 2±0 | 2±0 | 2±0 | 1±0 | 1±0 | 1±0 | 1±0 |  | 1±0 | 2±0 | 1±0 | 3±1 |
| CosFace | 2±0 | 2±0 | 2±0 | 2±0 | 2±0 | 2±0 | 2±0 | 2±0 | 2±0 | 1±0 | 1±0 |  | 2±0 | 1±0 | 3±1 |
| ArcFace | 2±0 | 2±0 | 2±0 | 2±0 | 2±0 | 2±0 | 2±0 | 2±0 | 2±0 | 2±0 | 2±0 | 2±0 |  | 1±0 | 3±1 |
| SoftTriple | 2±0 | 2±0 | 2±0 | 2±0 | 2±0 | 2±0 | 2±0 | 2±0 | 1±0 | 1±0 | 1±0 | 1±0 | 1±0 |  | 3±1 |
| None | 3±1 | 3±1 | 3±1 | 3±1 | 3±1 | 4±1 | 3±1 | 3±1 | 3±1 | 3±1 | 3±1 | 3±1 | 3±1 | 3±1 |  |

Table 4. Jensen-Shannon Divergence between all loss functions on the Cars196 dataset. All values are in percent.

|  | Ranking | | | | | | | | | Classification | | | | | |
|---|---|---|---|---|---|---|---|---|---|---|---|---|---|---|---|
|  | Contrastive | Triplet | NTXent | Margin | Margin/class | FastAP | SNR Con. | MS | MS+Miner | ProxyNCA | N. Softmax | CosFace | ArcFace | SoftTriple | None |
| Contrastive |  | 2±0 | 2±0 | 2±0 | 2±0 | 2±0 | 2±0 | 1±0 | 2±0 | 2±0 | 2±0 | 2±0 | 2±0 | 2±0 | 2±0 |
| Triplet | 2±0 |  | 1±0 | 1±0 | 1±0 | 2±0 | 2±0 | 1±0 | 1±0 | 1±0 | 1±0 | 2±0 | 1±0 | 2±0 | 2±0 |
| NTXent | 2±0 | 1±0 |  | 1±0 | 1±0 | 1±0 | 1±0 | 1±0 | 1±0 | 1±0 | 1±0 | 2±0 | 1±0 | 2±0 | 2±0 |
| Margin | 2±0 | 1±0 | 1±0 |  | 1±0 | 2±0 | 1±0 | 1±0 | 1±0 | 1±0 | 1±0 | 2±0 | 2±0 | 2±0 | 2±0 |
| Margin/class | 2±0 | 1±0 | 1±0 | 1±0 |  | 1±0 | 1±0 | 1±0 | 1±0 | 1±0 | 1±0 | 2±0 | 1±0 | 2±0 | 2±0 |
| FastAP | 2±0 | 2±0 | 1±0 | 2±0 | 1±0 |  | 1±0 | 1±0 | 1±0 | 2±0 | 1±0 | 2±0 | 2±0 | 2±0 | 2±1 |
| SNR Con. | 2±0 | 2±0 | 1±0 | 1±0 | 1±0 | 1±0 |  | 1±0 | 1±0 | 1±0 | 1±0 | 2±0 | 2±0 | 1±0 | 2±0 |
| MS | 1±0 | 1±0 | 1±0 | 1±0 | 1±0 | 1±0 | 1±0 |  | 1±0 | 1±0 | 1±0 | 1±0 | 1±0 | 1±0 | 2±0 |
| MS+Miner | 2±0 | 1±0 | 1±0 | 1±0 | 1±0 | 1±0 | 1±0 | 1±0 |  | 1±0 | 1±0 | 2±0 | 1±0 | 1±0 | 2±0 |
| ProxyNCA | 2±0 | 1±0 | 1±0 | 1±0 | 1±0 | 2±0 | 1±0 | 1±0 | 1±0 |  | 1±0 | 2±0 | 1±0 | 1±0 | 2±0 |
| N. Softmax | 2±0 | 1±0 | 1±0 | 1±0 | 1±0 | 1±0 | 1±0 | 1±0 | 1±0 | 1±0 |  | 2±0 | 1±0 | 1±0 | 2±0 |
| CosFace | 2±0 | 2±0 | 2±0 | 2±0 | 2±0 | 2±0 | 2±0 | 1±0 | 2±0 | 2±0 | 2±0 |  | 1±0 | 1±0 | 2±0 |
| ArcFace | 2±0 | 1±0 | 1±0 | 2±0 | 1±0 | 2±0 | 2±0 | 1±0 | 1±0 | 1±0 | 1±0 | 1±0 |  | 1±0 | 2±0 |
| SoftTriple | 2±0 | 2±0 | 2±0 | 2±0 | 2±0 | 2±0 | 1±0 | 1±0 | 1±0 | 1±0 | 1±0 | 1±0 | 1±0 |  | 2±0 |
| None | 2±0 | 2±0 | 2±0 | 2±0 | 2±0 | 2±1 | 2±0 | 2±0 | 2±0 | 2±0 | 2±0 | 2±0 | 2±0 | 2±0 |  |

Table 5. Jensen-Shannon Divergence between all loss functions on the CUB200 dataset. All values are in percent.

# B. Examples of the Qualitative Pixel Level Analysis

Figures 1 to 9 show saliency maps for example images from all three datasets Cars196, CUB200, and SOP.

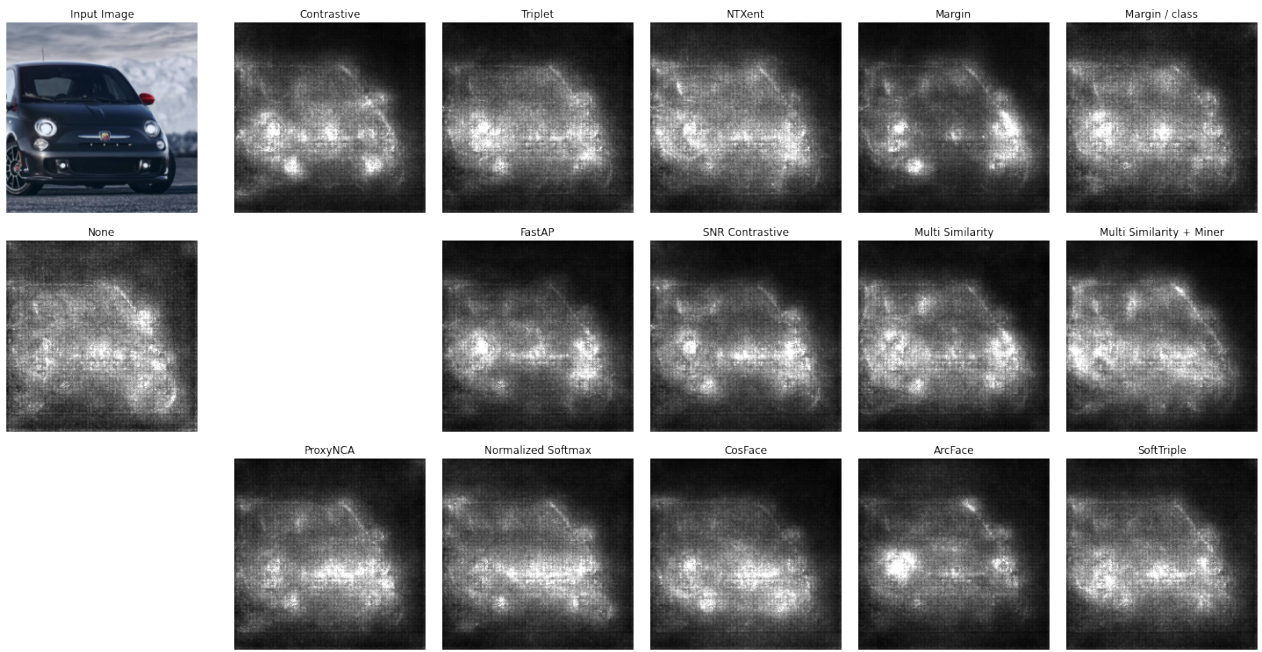

Figure 1. Saliency maps of a sample image from Cars196.

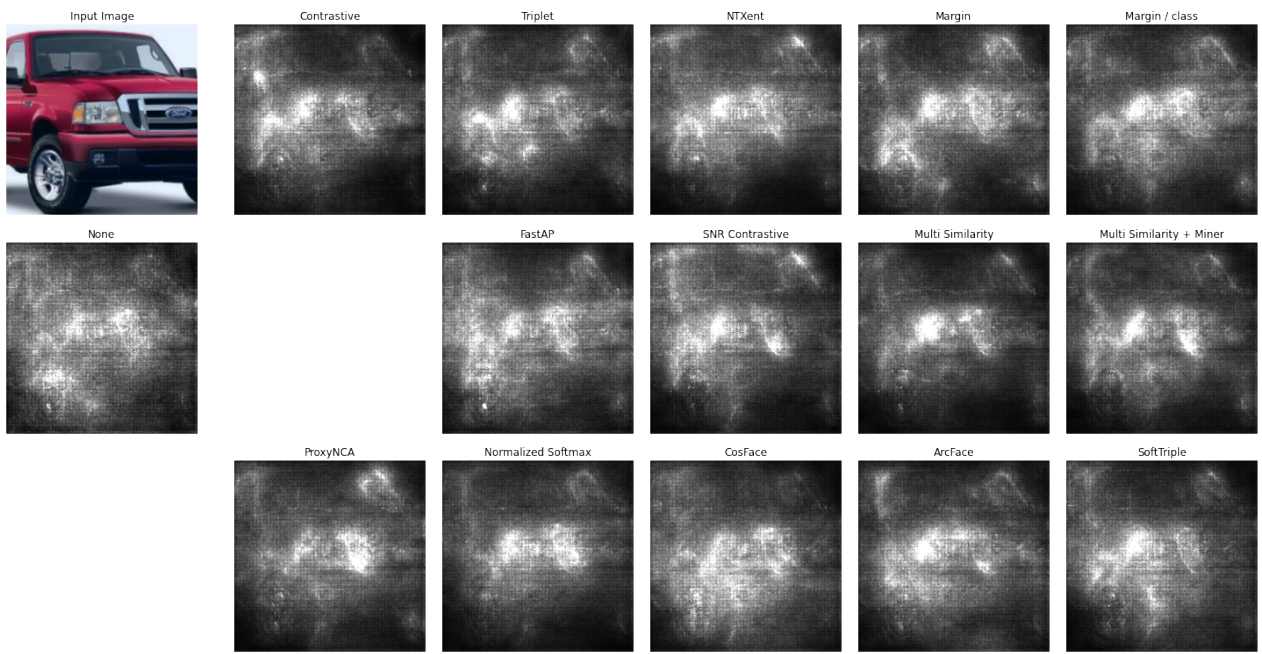

Figure 2. Saliency maps of a sample image from Cars196.

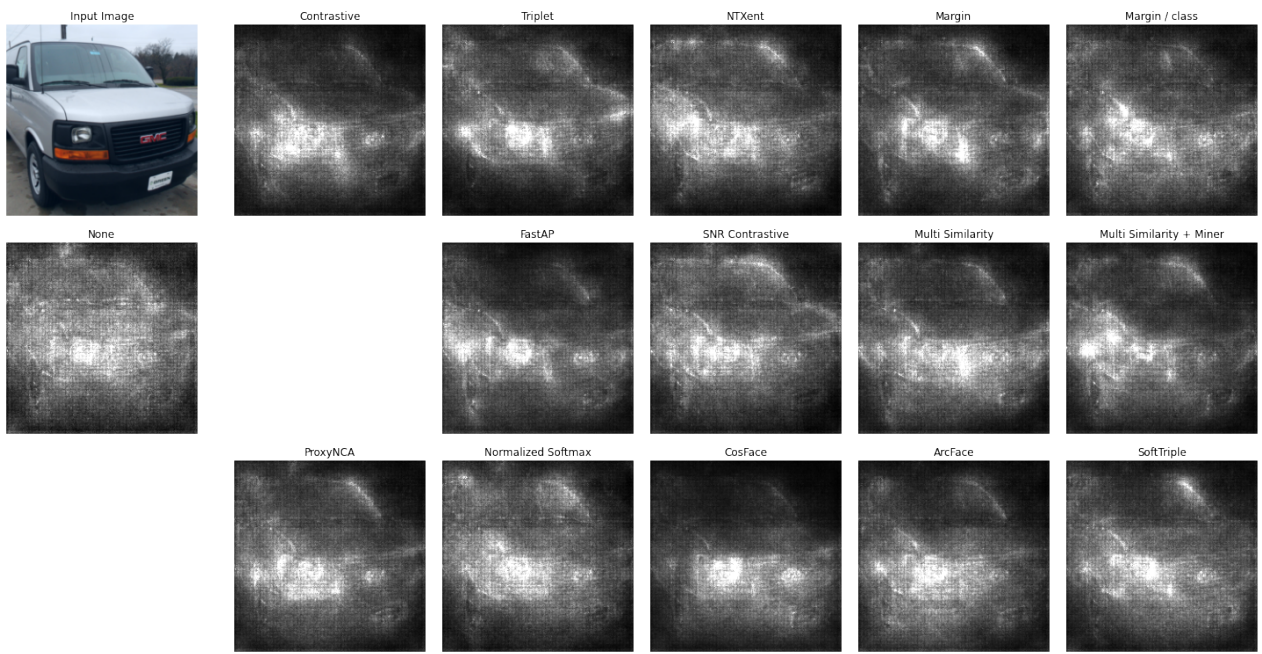

Figure 3. Saliency maps of a sample image from Cars196.

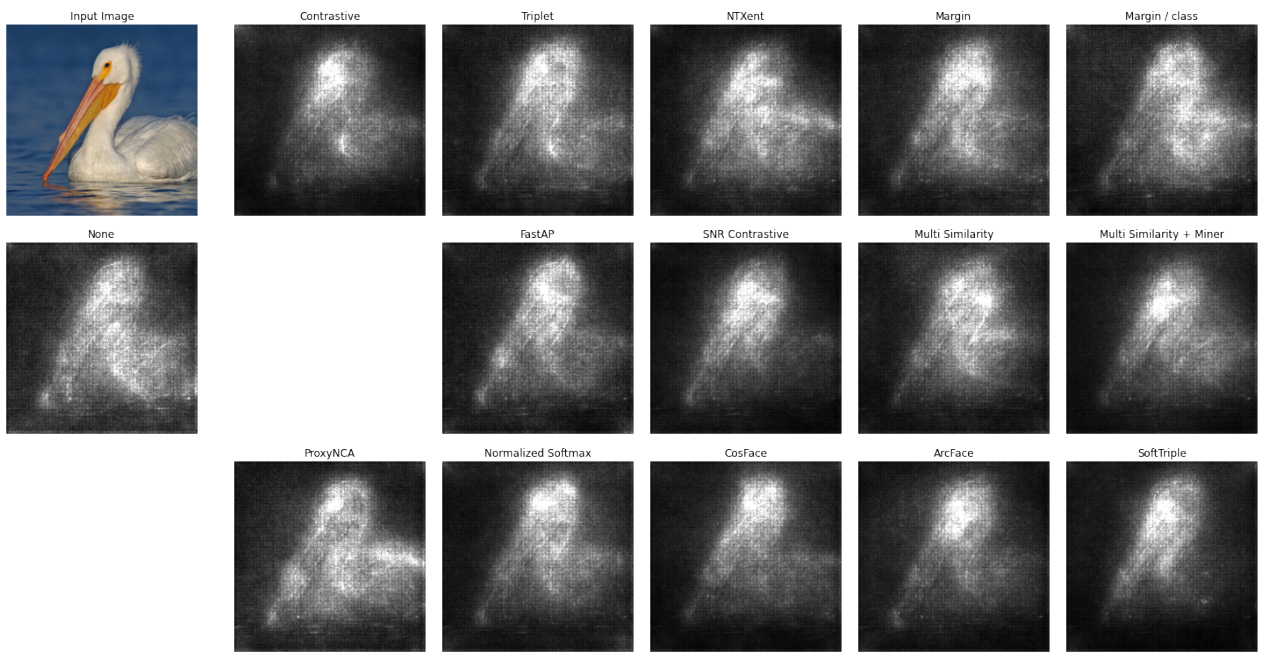

Figure 4. Saliency maps of a sample image from CUB200.

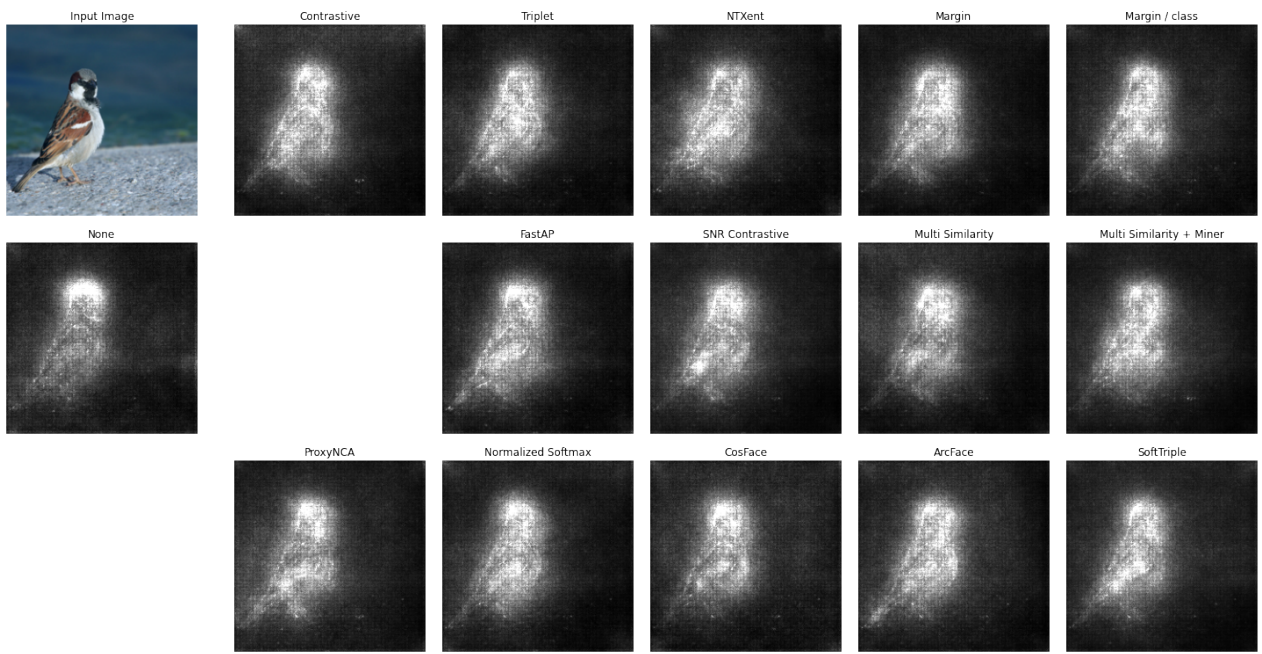

Figure 5. Saliency maps of a sample image from CUB200.

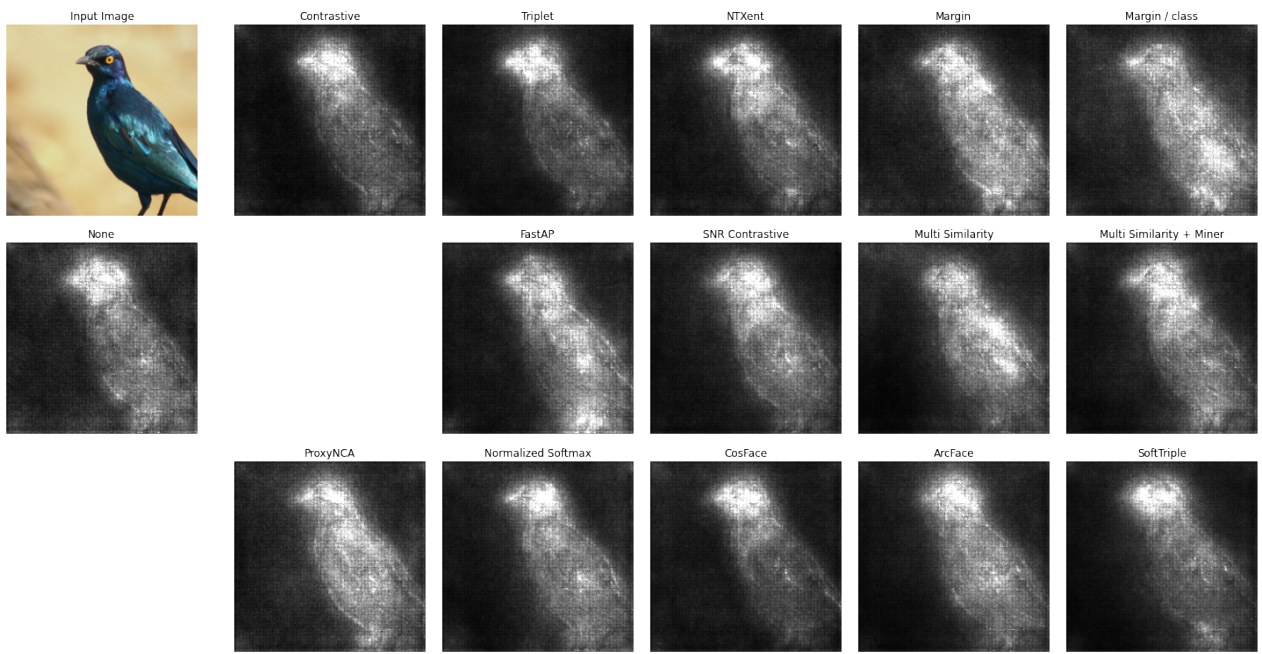

Figure 6. Saliency maps of a sample image from CUB200.

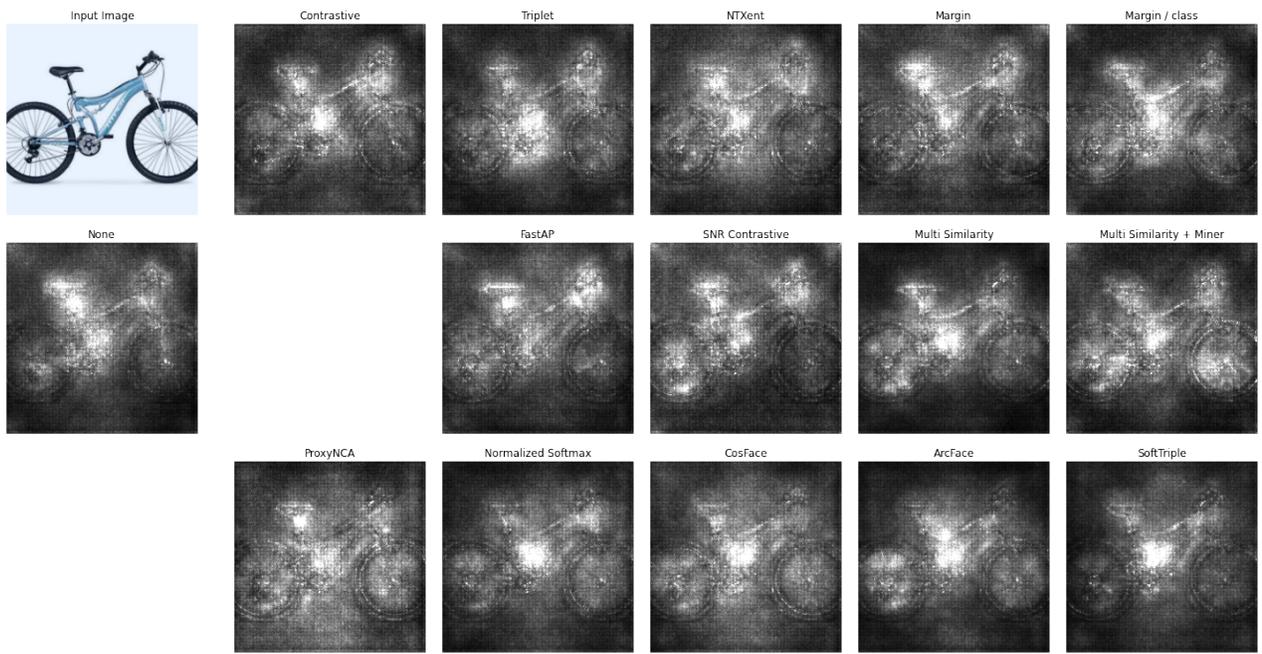

Figure 7. Saliency maps of a sample image from SOP.

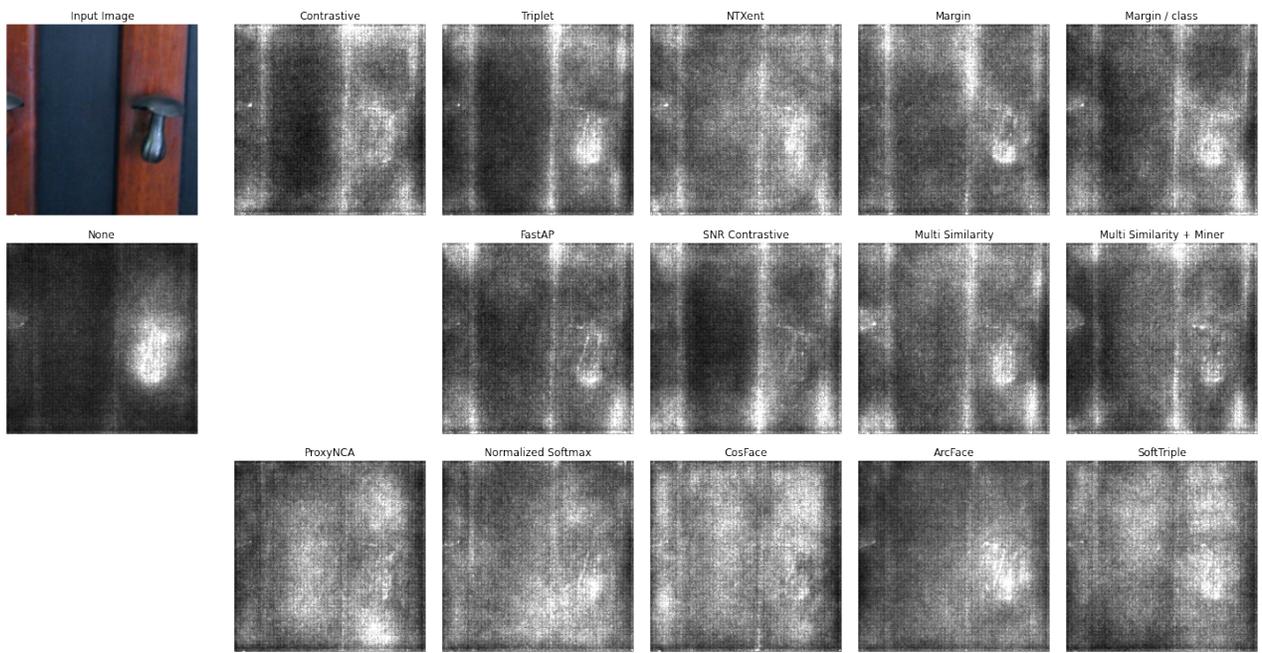

Figure 8. Saliency maps of a sample image from SOP.

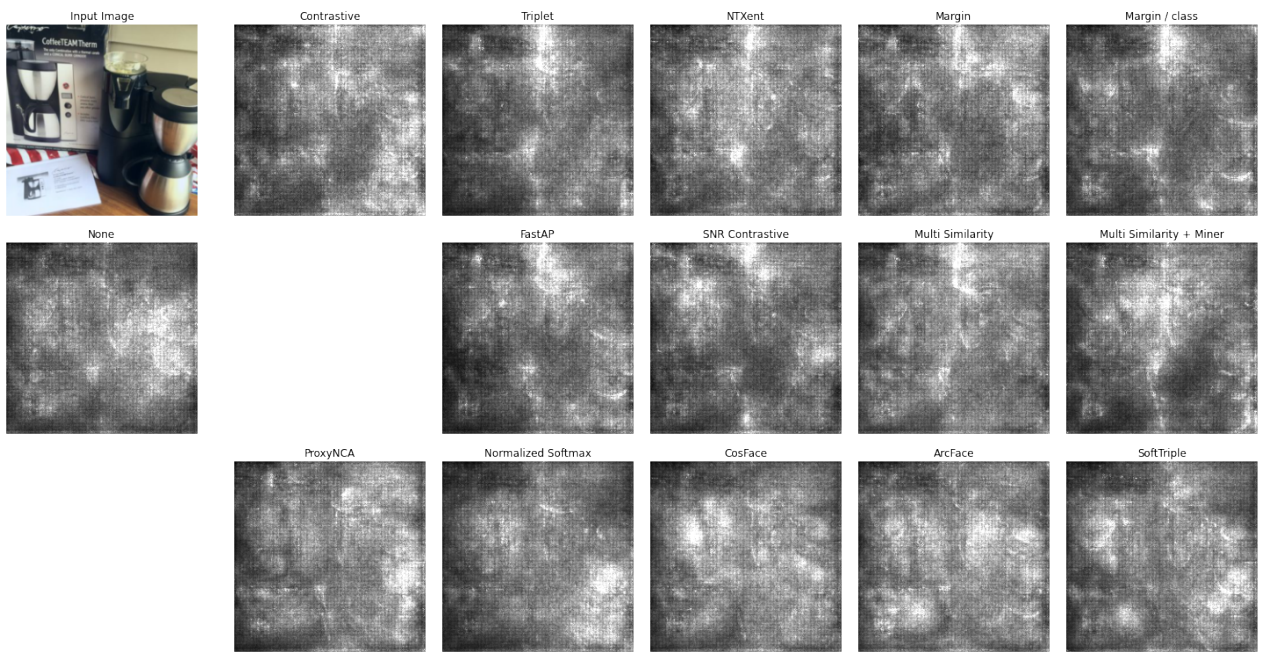

Figure 9. Saliency maps of a sample image from SOP.